\documentclass[journal]{IEEEtran}

\usepackage{url}
\usepackage{color}
\usepackage{amsmath}
\usepackage{array,graphicx,subfigure}
\usepackage{multirow}

\usepackage{algorithm}
\usepackage{algpseudocode}

\newcommand{\tabincell}[2]{\begin{tabular}{@{}#1@{}}#2\end{tabular}}

\newcommand{\cut}[1]{}

\newcommand{\etal}{\textit{et al}.}
\newcommand{\ie}{\textit{i}.\textit{e}.}
\newcommand{\eg}{\textit{e}.\textit{g}.}

\newcommand{\red}[1]{\textcolor{red}{#1}}
\newcommand{\blue}[1]{\textcolor{blue}{#1}}
\newcommand{\black}[1]{\textcolor{black}{#1}}

\ifCLASSINFOpdf
  
\else
 
\fi

\begin{document}

\title{Deep Self-Supervised Representation Learning  for Free-Hand Sketch}

\author{Peng~Xu,~Zeyu~Song,~Qiyue~Yin,~Yi-Zhe~Song,~\IEEEmembership{Senior~Member,~IEEE,}
        and~Liang~Wang,~\IEEEmembership{Fellow,~IEEE}
\thanks{Peng Xu is with School of Computer Science and Engineering, Nanyang Technological University, Singapore. E-mail: peng.xu@ntu.edu.sg~Homepage: http://www.pengxu.net/}
\thanks{Zeyu Song is with Beijing University of Posts and Telecommunications, Beijing, China. E-mail: szy2014@bupt.edu.cn }
\thanks{Qiyue Yin is with Institute of Automation, Chinese Academy of Sciences, Beijing, China. E-mail: qyyin@nlpr.ia.ac.cn }
\thanks{Yi-Zhe Song is with SketchX Lab, Centre for Vision, Speech and Signal
Processing (CVSSP), University of Surrey, United Kingdom. E-mail: y.song@surrey.ac.uk }
\thanks{Liang Wang is with Institute of Automation, Chinese Academy of Sciences, Beijing, China. E-mail: wangliang@nlpr.ia.ac.cn }
\thanks{This work was done bofore Peng Xu joined NTU.}
}


\maketitle

\begin{abstract}
In this paper, we tackle for the first time, the problem of self-supervised representation learning for free-hand sketches. This importantly addresses a common problem faced by the sketch community -- that annotated supervisory data are difficult to obtain. This problem is very challenging in that sketches are highly abstract and subject to different drawing styles, making existing solutions tailored for photos unsuitable. Key for the success of our self-supervised learning paradigm lies with our sketch-specific designs: (i) we propose a set of pretext tasks specifically designed for sketches that mimic different drawing styles, and (ii) we further exploit the use of a textual convolution network (TCN) in a dual-branch architecture for sketch feature learning, as means to accommodate the sequential stroke nature of sketches. We demonstrate the superiority of our sketch-specific designs through two sketch-related applications (retrieval and recognition) on a million-scale sketch dataset, and show that the proposed approach outperforms the state-of-the-art unsupervised representation learning methods, and significantly narrows the performance gap between with supervised representation learning.\footnote{PyTorch code of this work is available at \textcolor{blue}{\url{https://github.com/zzz1515151/self-supervised_learning_sketch}}.}
\end{abstract}


\IEEEpeerreviewmaketitle

\section{Introduction}

\IEEEPARstart{D}{eep} learning approaches have now delivered practical-level performances on various artificial intelligence tasks~\cite{song2017parameter,qin2018dsgan,tc3d2018liukun,guo2018hierarchical,qin2018robust,zhang2018image,wei2019adversarial,li2019pdr,xie2019soft,liu2019exploring,xu2019learning,chang2020mutualchannel,guo2020zero,qin2020generative}. However, most of the state-of-the-art deep models still rely on a massive amount of annotated supervisory data. These labor-intensive supervisions are so expensive that they have become a bottleneck of the general application of deep learning techniques. As a result, deep unsupervised representation learning~\cite{caron2018deep,gidaris2018unsupervised,kolesnikov2019revisiting} has gained considerable attention in recent days.

However, most of existing deep learning based unsupervised representation methods in computer vision area are engineered for photo~\cite{gidaris2018unsupervised} and video~\cite{kim2019self}. Unsupervised learning for sketches on the other hand remains relatively under-studied. It is nonetheless an important topic -- the lack of annotated data problem is particularly salient for sketches, since unlike photos that can be automatically crawled from the internet, sketches have to be drawn one by one.  


Sketch-related research has flourished in recent years~\cite{xu2020deep}, largely driven by the ubiquitous nature of touchscreen devices. Many problems have been studied to date, including
sketch recognition~\cite{schneider2014sketchFV,li2018sketch,xie2019deep,xu2019multigraph}, sketch hashing~\cite{xu2018sketchmate,liu2017deephashing}, sketch-based image retrieval~\cite{songjifei2017sketch,xu2018cross,dey2019doodle}, sketch synthesis~\cite{Chen_2018_CVPR}, segmentation~\cite{schneider2016sketchSegmentation}, scene understanding~\cite{ye2016human}, abstraction~\cite{muhammad2018learning}, just to name a few. However, almost all existing sketch-related deep learning techniques work in supervised settings, relying upon the manually labeled sketch datasets~\cite{sangkloy2016sketchy,songjifei2017sketch} collected via crowdsourcing. 

Despite decent performances reported, research progress on sketch understanding is largely bottlenecked by the size of annotated datasets (in their thousands). Recent efforts have since been made to create large-scale datasets ~\cite{ha2018sketchrnn,dey2019doodle}, yet their sizes and category coverage (annotated labels) are still far inferior to their photo counterparts~\cite{deng2009imagenet}. Furthermore, perhaps more importantly, sketch datasets suffer from being not easily extendable for sketches have to be manually produced other than automatically crawled from the internet. In this paper, we attempt to offer a new perspective to alleviate the data scarcity problem -- we move away from the commonly supervised learning paradigm, and for the first time study the novel and challenging problem of self-supervised representation learning for sketches.

\begin{figure*}[!t]
	\centering
		\includegraphics[width=\textwidth]{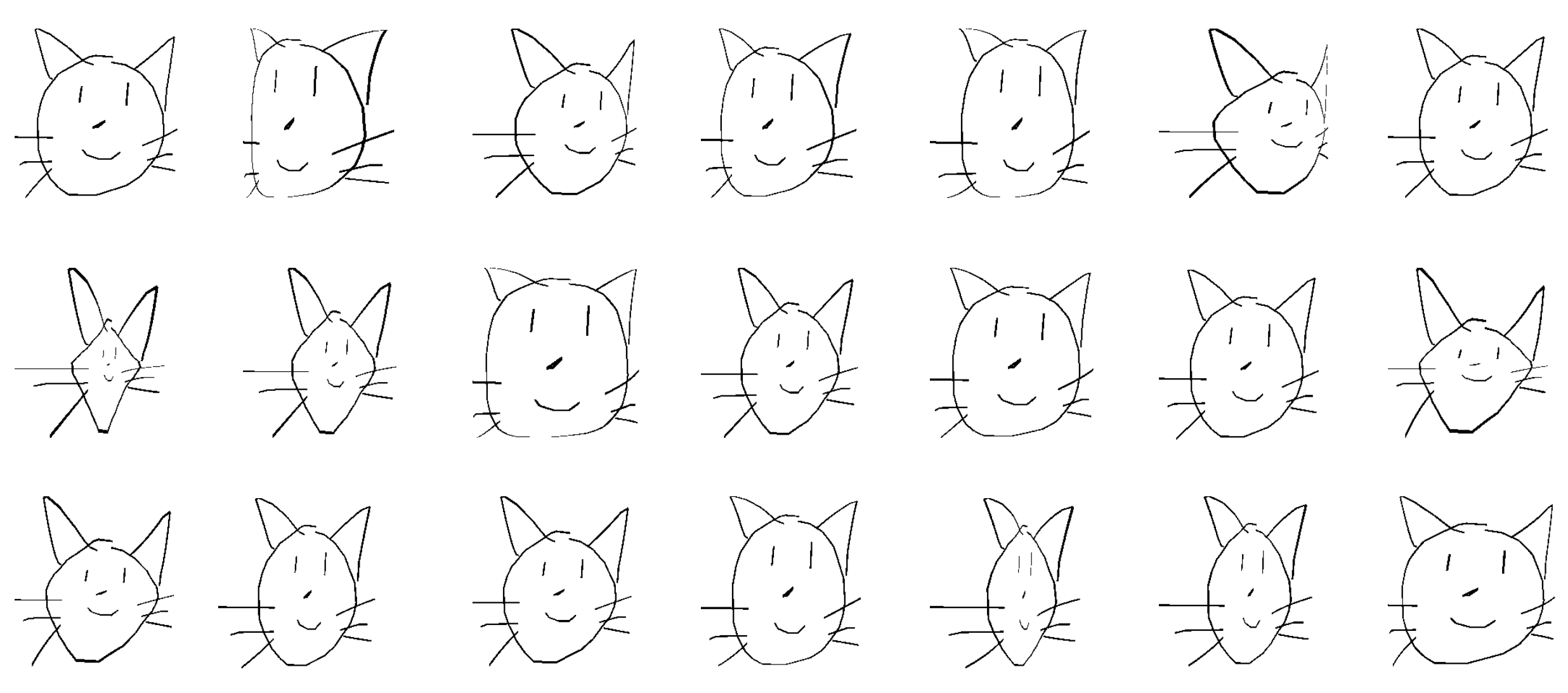}
	\caption{Illustration of the human sketching diversity. Different persons have different drawing styles.}
	\label{fig:cats}
\end{figure*}

\begin{figure}[!t]
	\centering
		\includegraphics[width=\columnwidth]{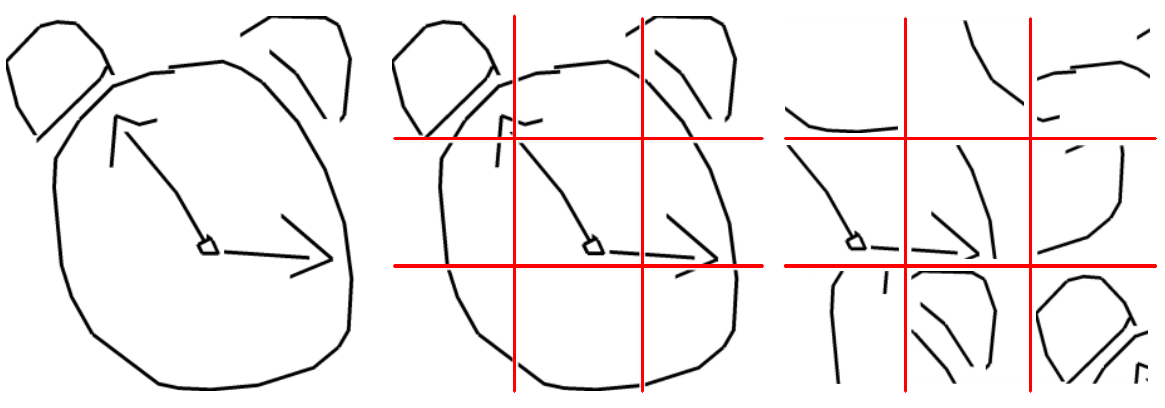}
	\caption{A sketch of alarm clock with its patches and the shuffled tiles. Sketch patches are too abstract to recognize, due to the stroke sparsity.}
	\label{fig:sketch_jigsaw}
\end{figure}

Solving the self-learning problem for sketches is however non-trivial. Sketches are distinctively different to photos -- a sketch is a temporal sequence of black and white strokes, whereas photos are collection of static pixels exhibiting rich color and texture information. Furthermore, sketches are also highly abstract, and subjected to different drawing styles. All such unique characteristics made self-supervised pretext tasks designed for photo fail to perform on sketch. This is mainly because the commonly adapted patch-based approaches~\cite{zhang2016colorful,doersch2015unsupervised,noroozi2016unsupervised} are not compatible with sketches -- sketches are formed of sparse strokes as oppose to dense pixel patches (see Figure~\ref{fig:sketch_jigsaw}). Our first contribution is therefore a set of sketch-specific pretext tasks that attempt to mimic the various drawing styles incurred in sketches. As shown in Figure~\ref{fig:cats}, human sketching styles are highly diverse. More specially, we first define a set of geometric deformations that simulate variations in human sketching (see Figure~\ref{fig:DT_samples}). Based on these deformations, we design a set of binary classification pretext tasks to train a deep model that estimates whether a geometric deformation has been applied to the input sketch. Intuitively, this encourages the model to learn to recognize sketches regardless of drawing styles (deformations), in turn forcing the model to learn to represent the input. This is akin to the intuition used by~\cite{gidaris2018unsupervised} by asking for rotation invariance but is otherwise specifically designed for sketches.


As our second contribution, we further exploit the use of a textual convolution network (TCN) to address the temporal stroke nature of sketches, and propose a dual-branch CNN-TCN architecture serving as feature extractor for sketch self-supervised representation learning. Current state-of-the-art feature extractors for sketches typically employ a RNN architecture to model stroke sequence~\cite{li2018sketch,sarvadevabhatla2016enabling}. However, a key insight highlighted by recent research, which we share in this paper, is that local stroke ordering can be noisy, and that invariance in sketching is achieved at stroke-group (part) level~\cite{muhammad2018learning}. This means that RNN-based approaches that model on stroke-level might work counter-productively. Using a TCN however, we are able to use different sizes of $1D$ convolution kernels to perceive strokes at different granularities (groups), hence producing more discriminative sketch features to help self-supervised learning.

Extensive experiments on million-scale sketches evaluate the superiority of our sketch-specific designs. In particular, we show that, for sketch retrieval and 
recognition, our self-supervised presentation approach not only outperforms existing state-of-the-art unsupervised representation learning techniques (\eg, self-supervised, clustering-based, reconstruction-based, GAN-based), but also significantly narrows the performance gap with supervised representation learning.

Our contributions can be summarized as: (i) Motivated by human free-hand drawing styles and habits, we propose a set of sketch-specific self-supervised pretext tasks within deep learning pipeline, which provide supervisory signals for semantic feature learning. 
(ii) We for the first time exploit textual convolution network (TCN) for sketch feature modeling, and propose a dual-branch CNN-TCN feature extractor architecture for self-supervised sketch representation learning. 
We also show that our CNN-TCN architecture generalizes well under supervised feature representation settings (\eg, fully supervised sketch recognition), and outperforms existing architectures serving for sketch feature extraction (\ie, CNN, RNN, CNN-RNN).

The rest of this paper is organized as follows:
Section~\ref{sec:relatedwork} briefly summarizes related work.
Section~\ref{sec:methodology} describes our proposed sketch-specific self-supervised representation learning method, introducing our proposed sketch-specific self-supervised pretext tasks and feature extractor architecture.
Experimental results and discussion are presented in Section~\ref{sec:experiments}.
We present some conclusions and insights in Section~\ref{sec:conclusion}.
Finally, future work is discussed in Section~\ref{sec:future_work}.

\section{Related Work}
\label{sec:relatedwork}
\subsection{Self-Supervised Pretext Tasks}
Deep learning based unsupervised representation learning techniques can be broadly categories into self-supervised methods~\cite{gidaris2018unsupervised}, auto-encoder~\cite{pathak2016context},
generative adversarial network~\cite{donahue2016adversarial,radford2015unsupervised}, and  clustering~\cite{dosovitskiy2014discriminative,liao2016learning}.
Currently, self-supervised representation learning~\cite{gidaris2018unsupervised} has achieved state-of-the-art performance on computer vision tasks such as classification, and segmentation.
The key technique in self-supervised approaches is defining pretext tasks to force the model learn how to represent the feature of input. 
Existing self-supervised pretext tasks mainly include patch-level predictions~\cite{doersch2015unsupervised,mundhenk2018improvements,noroozi2018boosting}, and 
~image-level predictions~\cite{gidaris2018unsupervised}. Sketch is essentially different from photo, thus patch-based pretext tasks (\eg, predicting relative position of image patches~\cite{doersch2015unsupervised}) fail to work on sketch, due to that sketch patches are too abstract to recognize.
The colorization-related pretext tasks~\cite{zhang2016colorful,kim2018learning
} are also unsuitable for sketch since sketches are color-free. To the best of our knowledge, our work is the first self-supervised representation learning work on sketches.

\subsection{Sketch Feature Extractor Architecture}
Most prior works model sketch as static pixel picture and use CNN as feature extractor, neglecting the sequential drawing patterns on stroke level. 
\cite{ha2018sketchrnn} proposed the groundbreaking work that uses RNN to model strokes, which was the first to explore the temporal nature of sketch. 
\cite{xu2018sketchmate} proposed a dual-branch CNN-RNN architecture,
using CNN to extract abstract visual concepts and RNN to learn sketching temporal orders.
Some tandem architectures also have been proposed, including CNN on the top of RNN~\cite{li2018sketch}, RNN on the top of CNN~\cite{sarvadevabhatla2016enabling}.
However, all previous sketch feature extractor architectures have been proposed in fully-supervised settings.
Moreover, as stated in~\cite{kolesnikov2019revisiting}, standard network architecture design recipes do not necessarily translate from the fully-supervised setting to the self-supervised setting. Therefore, in this paper, we explore a novel feature extractor architecture specifically purposed for sketch self-supervised learning. The TCN architecture also appropriately address the temporal nature of sketches, while accommodating for stroke granularity. To the best of our knowledge, this work is the first probe that models sketch feature using TCN.

\section{Sketch-Specific Self-Supervised Representation Learning}
\label{sec:methodology}
\subsection{Problem Formulation}
We assume training dataset $X$ in the form of $N$ sketch samples: $X=\{X_i=(P_i,S_i)\}^N_{i=1}$. 
Each sketch sample consists of a sketch picture $P_i$ 
and a corresponding sketch stroke sequence $S_i$. 
We aim to learn semantic feature $\mathcal{F}(P_i, S_i)$ for sketch sample $X_i$ in a self-supervised manner, in which $\mathcal{F}$ denotes feature extraction.

\subsection{Overview}
We aim to extract semantic features for free-hand sketch in self-supervised approach.
\textcolor{black}{Inspired by} the state-of-the-art self-supervised method~\cite{gidaris2018unsupervised}, 
we try to train a deep model to estimate the geometric deformation applied to the original input, hoping the model is able to learn how to capture the features of the input. 
Thus, we would define a set of $L$ discrete geometric deformations $\mathcal{T} = \{\mathcal{T}_{\ell}(\cdot)\}_{\ell=1}^{L}$. In our self-supervised setting, given a sketch sample $X_i$, we do not know its class label, but we can generate some deformed samples by applying our deformation operators on it as
\begin{equation}
\label{equ:transformation}
\begin{split}
X_i^{\ell} = \mathcal{T}_{\ell}(X_i),
\end{split}
\end{equation}
where $\ell \in [1, L]$ denotes the label of deformation.
 Therefore, given a training sample $X_i$, the output of the deep model can be formulated as $L$-way softmax, which can be denoted as ${\mathcal{F}_{\Theta}^{logits}(X_i)}$, assuming that $\Theta$ indicates deep neural network parameters. 
Given training dataset, our objective is to minimize the cross entropy loss over $L$-way softmax:
\begin{equation}
\label{equ:objective}
\begin{split}
\min \limits_{\Theta} \frac{1}{N \times L} \sum_{i=1}^{N} \sum_{\ell=1}^{L} - \log \frac{\mathrm{e}^{\mathcal{F}_{\Theta}^{logits, {\ell}}(X_i^{\ell})}}{\sum_{\widehat{\ell}=1}^{L} \mathrm{e}^{\mathcal{F}_{\Theta}^{logits, \widehat{\ell}}(X_i^{\ell})}},
\end{split}
\end{equation}
where 
${\mathcal{F}_{\Theta}^{logits, \widehat{\ell}}(X_i^{\ell})}$ indicates the $\widehat{\ell}$th value of the output probability logits for deformed sample~$X_i^{\ell}$.

Based on above analysis, we \textcolor{black}{next} need to 
find 
geometric transformations to define pretext classification tasks that can provide useful supervision signal to drive the model to capture feature of sketch.
Sketch can be formated as picture in pixel space, so that the rotation-based self-supervised technique also can be applied to it.
However, sketch has several intrinsic traits, \eg, (i) Sketch is highly abstract. (ii) Sketch can be formatted as a stroke sequence.
In the following, we propose a set of sketch-specific self-supervised pretext tasks 
and a novel sketch-specific feature extractor architecture.
\subsection{Sketch-Specific Self-Supervised Pretext Tasks}
Free-hand sketch is a special form of visual data sharing some similarities with handwritten character.
Even if one person draw the same object or scene more than once, the obtained sketches are impossible to be completely the same. Moreover, different persons have different drawing styles, habits, and abilities. If ask several persons to draw the same cat, some persons maybe habitually draw it with slim style while some persons maybe draw it as a fat cat.
Although there are large variations among the obtained sketchy cats under different drawing styles, the basic topological structures of them are the same.
\textcolor{black}{Inspired by} Handwritten Character Shape Correction~\cite{jinlianwen2000}, in this paper,
we aim to use \textbf{nonlinear functions} to model these 
flexible and irregular drawing deformations \textcolor{black}{, to define a set of discrete self-supervised pretext tasks for sketch. That is to train the deep model to judge whether the input has undergone one kind of deformations, hoping that the model is able to learn how to capture features of sketch.} 

\begin{figure}[!t]
	\centering
		\includegraphics[width=0.5\textwidth]{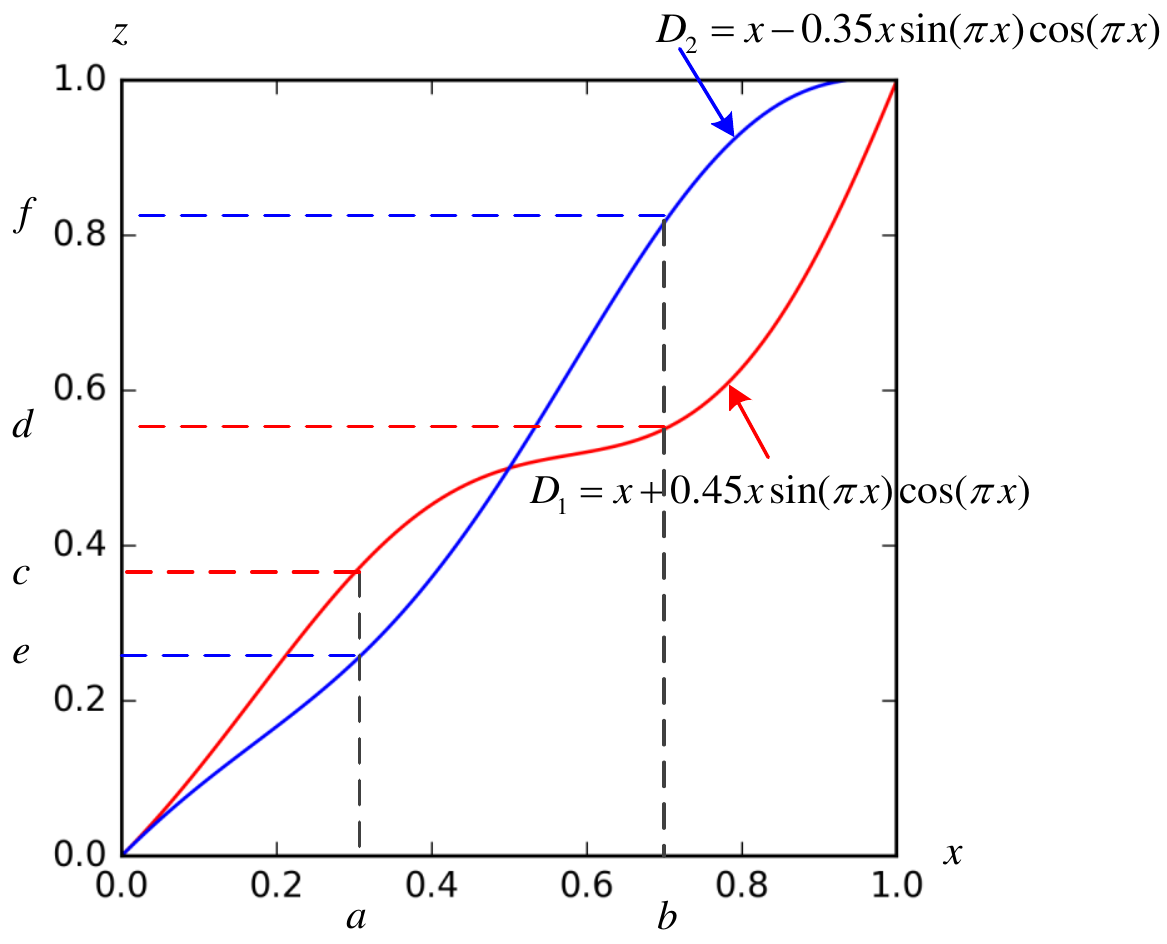}

   \caption{Illustration of two deformation curves with different deforming parameters.}
	\label{fig:dt_curves}
\end{figure}

Given a binaryzation sketch $X_i$ (stroke width is one pixel), its stroke sequence $S_i$ can be denoted as a series of coordinates of the black pixels:
\begin{equation}
\label{equ:S_i}
\begin{split}
S_i = \{(x_{k}, y_{k})\}_{k = 1}^{K_i},
\end{split}
\end{equation}
where $(x_{k}, y_{k})$ is the $k$th black pixel (point) of $X_i$, and $K_i$ is the total amount of black pixels of $X_i$. 
We can normalize the coordinates for each black pixel such that $x_{k}, y_{k} \in [0, 1]$.
\textcolor{black}{Intuitively}, we can use arbitrary functions as displacement functions, and the horizontal and vertical directions are independent to each other. 
Therefore, the 
\textcolor{black}{deformable transformations} on \verb+x+ and \verb+y+ directions (denoted as $D_{x}(\cdot)$, $D_{y}(\cdot)$) for $X_i$ can be performed as
\begin{equation}
\left\{
\begin{aligned}
D_{x}(x_{k}) & =  x_{k} + f_{x}(x_{k}), & & k \in [1, K_i],\\
D_{y}(y_{k}) & =  y_{k} + f_{y}(y_{k}), & & k \in [1, K_i],
\end{aligned}
\right.
\end{equation}
where $f_{x}(\cdot)$ and $f_{y}(\cdot)$ are the \textcolor{black}{displacement functions} on \verb+x+ and \verb+y+ directions, respectively.
As stated in~\cite{jinlianwen2000}, deformation should meet some properties: (i) Displacement functions are nonlinear functions. (ii) Displacement functions are continuous and satisfy boundary conditions: $f_{x}(0, y) = f_{x}(1, y) = f_{y}(x, 0) = f_{y}(x, 1) = 0$. (iii) Displacement functions should be monotonically increasing functions so that deformation transformation could preserve the topology structure of the original sketch. (iv) Deformation should preserve the smoothness and connectivity of the original sketch.
Based on these constraints, we can use following function as one displacement function:
\begin{equation}
\label{equ:original_dt_function}
\left\{
\begin{aligned}
f(x) & =  \eta x [\sin(\pi \beta x + \alpha) \cos(\pi \beta x + \alpha) + \gamma],\\
f(0) & =  0,\\
f(1) & =  0,
\end{aligned}
\right.
\end{equation}
where $f(1) = 0$ derives $\gamma = - \sin(\pi \beta + \alpha) \cos(\pi \beta + \alpha)$. Thus this displacement function can be simplified as:
\begin{equation}
\label{equ:brief_dt_function}
\begin{split}
f(x) = \eta x [\sin(\pi \beta x + \alpha) \cos(\pi \beta x + \alpha) \\ 
- \sin(\pi \beta + \alpha) \cos(\pi \beta + \alpha)], 
\end{split}
\end{equation}
where $\alpha$, $\beta$, 
and $\eta$ are constants.
$\eta$ controls the nonlinear mapping intensity. 
Here, we would take a concrete example
to illustrate how this trigonometric function can perform nonlinear deformation on sketch \textcolor{black}{picture}. 
If we set $\alpha = 0$, $\beta = 1$
for \eqref{equ:brief_dt_function}, the displacement function becomes as
\begin{equation}
\label{equ:example_dt_function}
f(x) = \eta x \sin(\pi x) \cos(\pi x).
\end{equation}
We can plot two deformation curves $D_1 = x + 0.45x\sin(\pi x) \cos(\pi x)$ and $D_2 = x - 0.35x\sin(\pi x) \cos(\pi x)$ in Figure~\ref{fig:dt_curves}.
We can observe that displacement function~\eqref{equ:example_dt_function} is nonlinear function that can map linear domain of x-axis into nonlinear domain of z-axis: (i) $D_1$ compresses $[a, b]$ into $[c, d]$, (ii) $D_2$ expands $[a, b]$ into $[e, f]$. 
As illustrated in Figure~\ref{fig:dt_curves}, various of nonlinear deformation effects can be obtained by selecting different regions of deformations and deforming parameter $\eta$. Uneven stroke widths are caused by interpolations during the deformations.
\begin{figure*}[!t]
	\centering
		\includegraphics[width=1.0\textwidth]{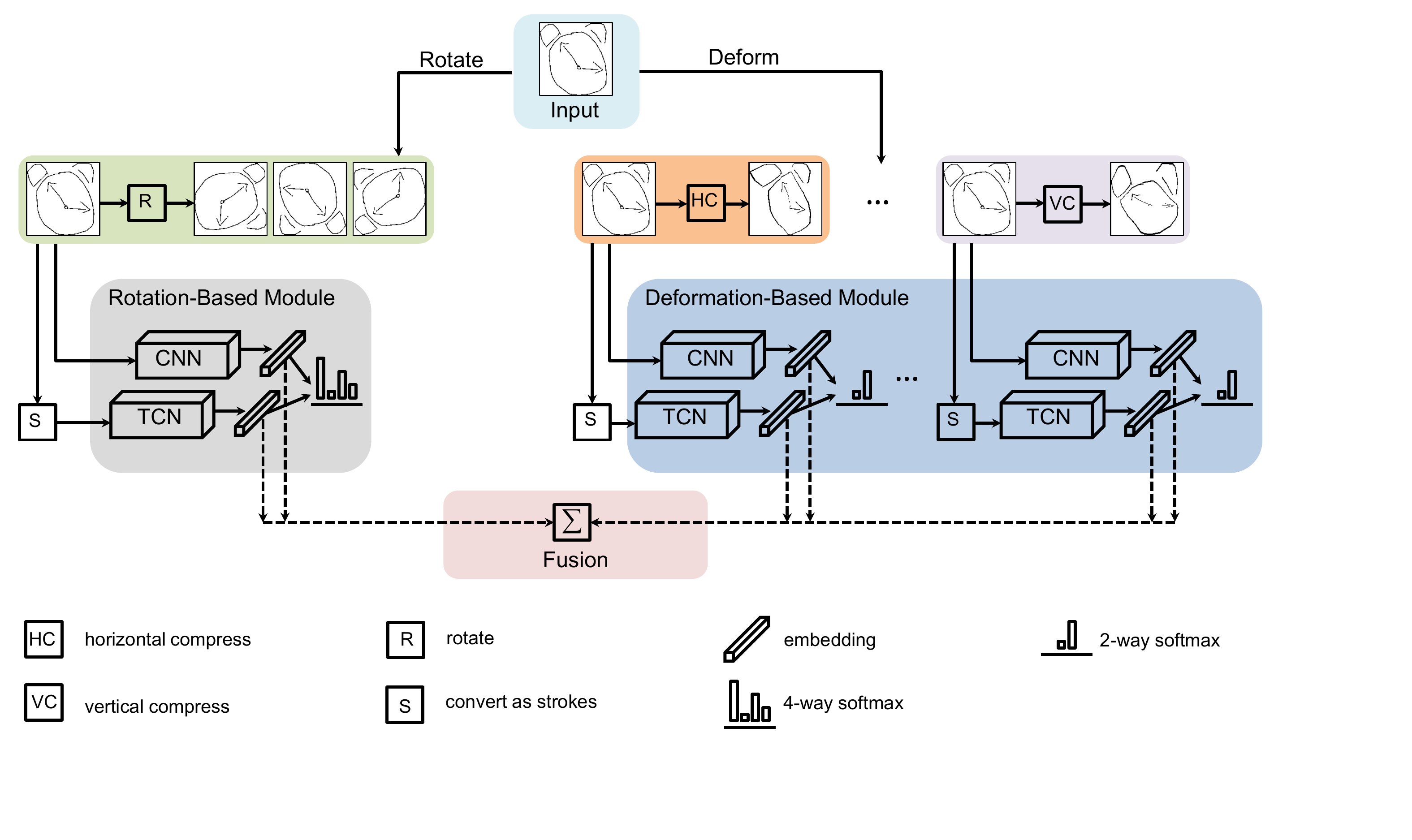}
   \caption{Illustration of our proposed sketch-specific self-supervised representation learning framework.}
	\label{fig:pipeline}
\end{figure*}


\begin{figure*}[!t]
	\centering
		\includegraphics[width=\textwidth]{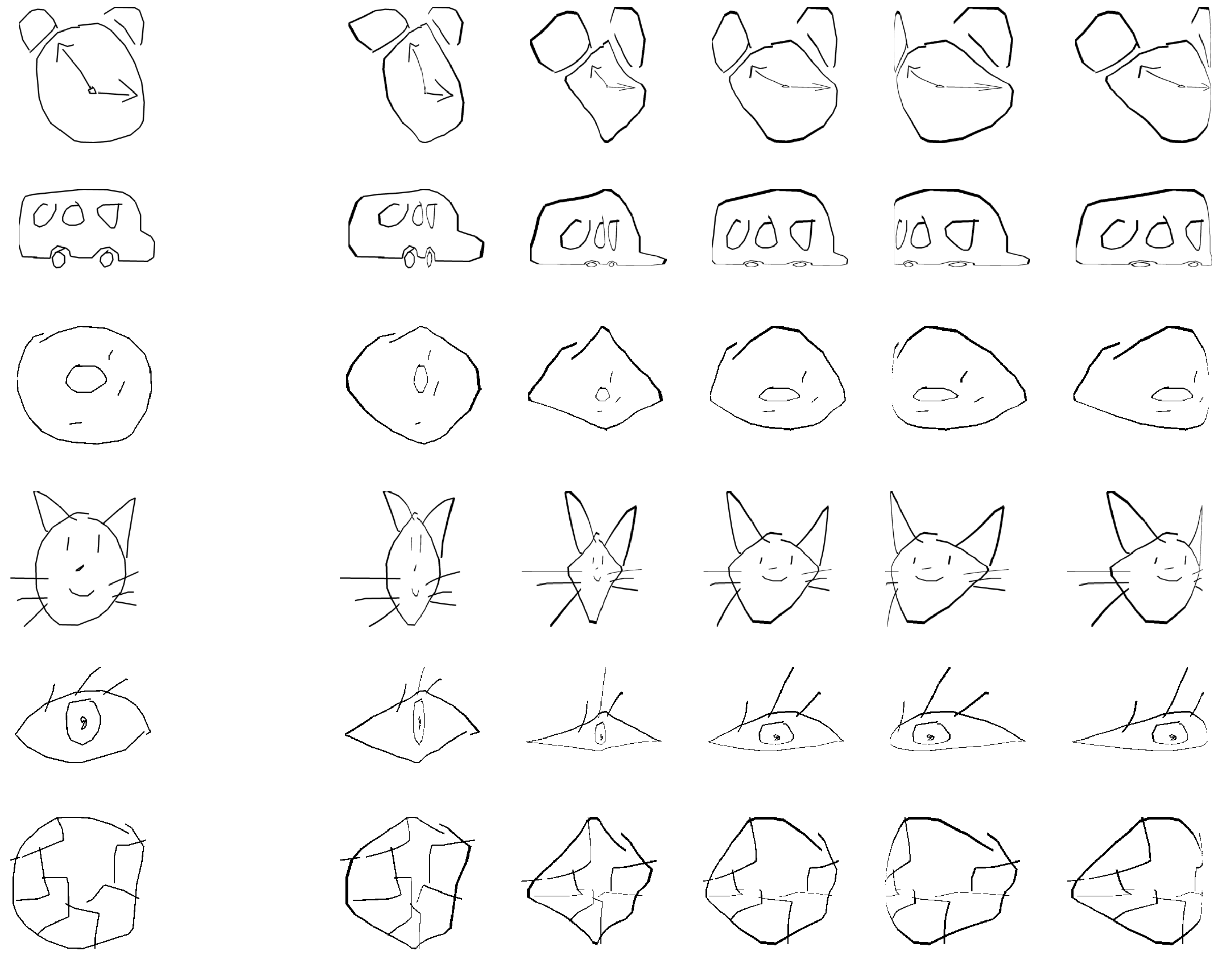}
	\caption{Illustration of six sketch samples (alarm clock, school bus, donut, cat, eye, soccer) and their deformation pictures. For each sample, original sketch is shown in the left column, while sketches presented in 2nd to 6th columns are respectively obtained by horizontal compression~(HC), centripetal compression~(CC), vertical compression~(VC), leftward compression~(LC), and rightward compression~(RC).}
	\label{fig:DT_samples}
\end{figure*}
Assuming that we aim to apply the nonlinear mapping of region $[a, b]$ of \eqref{equ:brief_dt_function} to a coordinate normalized sketch,
we can set
\begin{equation}
\left\{
\begin{aligned}
\pi \beta x + \alpha \big| _{x = 0}& = a,\\
\pi \beta x + \alpha \big| _{x = 1}& = b,
\end{aligned}
\right.
\end{equation}
such that
\begin{equation}
\label{equ:variable_substitute_2}
\begin{split}
\alpha = a, \beta = (b - a) / \pi.
\end{split}
\end{equation}
Taking \eqref{equ:variable_substitute_2} into \eqref{equ:brief_dt_function}, the 
deformations can be defined as: 
\begin{equation}
\label{equ:shape_correction_definition}
\left\{
\begin{aligned}
D(x_{k}) & = x_{k} + \eta_{1} x_{k} [\sin[(b_{1} - a_{1}) x_{k} + a_{1}] \\ & \times \cos[(b_{1} - a_{1}) x_{k} + a_{1}] - \sin b_{1} \cos b_{1}],\\
D(y_{k}) & = y_{k} + \eta_{2} y_{k} [\sin[(b_{2} - a_{2}) y_{k} + a_{2}] \\ & \times \cos[(b_{2} - a_{2}) y_{k} + a_{2}] - \sin b_{2} \cos b_{2}],
\end{aligned}
\right.
\end{equation}
where $a_{1}$, $b_1$, $a_2$, $b_2$, $\eta_{1}$, and $\eta_{2}$ are hyper-parameters,
conforming $0 \leq a_{1} \textless b_{1} \leq 1$ and $0 \leq a_{2} \textless b_{2} \leq 1$.
As shown in Figure~\ref{fig:DT_samples}, we can adjust $a$, $b$, $\eta$ to achieve different deformation effects to simulate different kinds of free-hand drawing habits.~(Limited by page space, only five kinds are shown.) In Figure~\ref{fig:DT_samples}, for each sketch sample, the left column is the original picture,  and the five columns on right are respectively transformed by horizontal compression~(HC), \textcolor{black}{centripetal} compression~(CC), vertical compression~(VC), leftward compression~(LC), and rightward compression~(RC).
Please note that nonlinear function based deformation can be applied to both sketch picture and sketch stroke sequence, while deformation upon sketch picture is convenient for visualization.

In this paper, we aim to define a set of sketch-specific binary classification taks that make the deep model to judge whether the input has undergone one kind of deformation, as sketch-specific self-supervised representation learning pretext tasks, hoping that the model is able to learn how to capture features of sketch.

\subsection{Sketch-Specific Feature Extractor Architecture}
Recently, Kolesnikov \etal ~\cite{kolesnikov2019revisiting} demonstrate that standard network architecture design recipes do not necessarily translate from fully-supervised setting to self-supervised setting. 
\textcolor{black}{We will also present similar phenomenon in following experiments that RNN-based feature extractor network fails to converge under our self-supervised training, while RNN-based networks have achieved satisfactory feature representation effects in previous supervised settings~\cite{xu2018sketchmate,ha2018sketchrnn}.}
Therefore, it is necessary to explore novel feature extractor network architecture upon sketch self-supervised setting.
Most of the previous sketch-related research works use CNNs as feature extractor.
In this work, we propose a dual-branch CNN-TCN architecture for sketch feature representation, utilizing CNN to extract abstract semantic concepts from $2D$ static pixel space and TCN to 
sequentially probe along sketch strokes by $1D$ convolution operation.
In particular, 
for sketch feature extraction,
the major advantage of TCN is that sequentially probing along sketch strokes at different granularities by varying its $1D$ convolution kernel sizes~(receptive fields). That is using small and large kernels to capture the patterns of short and long strokes, respectively. 

\subsection{Sketch-Specific Self-Supervised Representation Learning Framework}
Our proposed framework 
is illustrated in Figure~\ref{fig:pipeline}, containing two major components: rotation-based 
representation module
and deformation-based representation module. 
Quaternary classification on rotations~($0^\circ,90^\circ,180^\circ,270^\circ$) are used as pretext task to train the rotation-based representation module.
Our deformation-based representation module is extensible, which can consist of more than one representation sub-module.
For each deformation-based representation sub-module, we choose a specific nonlinear deformation, and train sub-module to estimate whether the chosen nonlinear deformation has been applied to the input sketch.
This is to say that the pretext task for each deformation-based representation sub-module is a binary classification.
\black{We empirically find that multiple binary classification based representation sub-modules work better than the single multi-class classification based representation module. This is mainly due to that classification of diverse deformations is difficult to be trained on the highly abstract sketches.} 
In our rotation-based representation module and deformation-based representation sub-modules, dual-branch CNN-TCN network serves as the feature extractor.
\cut{Generally, CNN and TCN capture complementary characteristics of sketches, i.e., spatial and temporal patterns, we concatenate their embeddings as the final self-supervised representation.}
\black{In this paper, we focus on developing a general framework for sketch self-supervised learning, thus over-complicated fusion strategies are not discussed here and will be thoroughly compared in the future work.
Moreover, CNN and TCN are essentially heterogeneous architectures, so that it's unpractical to train them synchronously.
Therefore, we train our CNNs and TCNs separately.} 
The detailed training and optimization are described in Algorithm~\ref{alg:1}.

During testing, given a sketch sample $(P_i, S_i)$, its feature representation can be defined as    
\begin{equation}
\label{equ:feature_fusion}
\begin{split}
\mathcal{F}(P_i, S_i) = \lambda^{r} \mathcal{F}_{{\Theta}}^{r}(P_i, S_i) + \sum_{j=1}^{J} \lambda^{d_j} \mathcal{F}_{{\Theta}}^{d_j}(P_i, S_i),
\end{split}
\end{equation}
where 
$\mathcal{F}_{{\Theta}}^{r}(\cdot)$~and~$\mathcal{F}_{{\Theta}}^{d_j}(\cdot)$
denote the feature extractions of 
rotation-based module
and the $j$th 
deformation-based sub-module
, respectively. $\lambda^{r}$~and~$\lambda^{d_j}$ are weighting factors.
The output feature of the rotation-based module is fused via
\begin{equation}
\label{equ:rotation_feature_fusion}
\begin{split}
\mathcal{F}^{r}(P_i, S_i) = {\mu^r} \mathcal{F}_{{\Theta}_c}^{r}(P_i) + (1 - {\mu^r}) \mathcal{F}_{{\Theta}_t}^{r}(S_i),
\end{split}
\end{equation}
where $\mathcal{F}_{{\Theta}_{c}}^{r}(\cdot)$ and $\mathcal{F}_{{\Theta}_{t}}^{r}(\cdot)$
denote the feature extractions of the CNN and TCN branches of rotation-based module, respectively. 
${\mu^r}$ is a weighting factor.
 Similarly, the output feature of the $j$th 
deformation-based sub-module
 is fused via
\begin{equation}
\label{equ:shape_correction_feature_fusion}
\begin{split}
\mathcal{F}_{{\Theta}}^{d_j}(P_i, S_i) = {\mu^{d_j}} \mathcal{F}_{{\Theta}_c}^{d_j}(P_i) + (1 - {\mu^{d_j}}) \mathcal{F}_{{\Theta}_t}^{d_j}(S_i),
\end{split}
\end{equation}
where 
$\mathcal{F}_{{\Theta}_c}^{d_j}(\cdot)$~and~$\mathcal{F}_{{\Theta}_t}^{d_j}(\cdot)$
indicate the feature extractions of the CNN and TCN branches of the $j$th
deformation-based sub-module
, respectively. ${\mu^{s_j}}$ is a weighting factor.

\begin{algorithm}[!t]
    	\caption{Learning algorithm for our sketch-specific self-supervised representation learning framework.}
  	\label{alg:1}
        \begin{algorithmic}
        \Require $X=\{X_i=(P_i,S_i)\}^N_{i=1}$.
        \State 1. Train rotation-based module $\mathcal{F}_{{\Theta}}^{r}(\cdot)$.
		  \State 1.1. Train CNN branch on $\{P_i\}^N_{i=1}$, and obtain $ \mathcal{F}_{{\Theta}_c}^{r}$. 
        \State 1.2. Train TCN branch on $\{S_i\}^N_{i=1}$, and obtain $ \mathcal{F}_{{\Theta}_t}^{r}$.
        \State 2. Train deformation-based module as following loop.
        \For{Each deformation-based sub-module $\mathcal{F}_{{\Theta}}^{d_j}(\cdot)$}
            \State 2.1 Train CNN branch on $\{P_i\}^N_{i=1}$, and obtain $\mathcal{F}_{{\Theta}_c}^{d_j}$.
            \State 2.2 Train TCN branch on $\{S_i\}^N_{i=1}$, and obtain $\mathcal{F}_{{\Theta}_t}^{d_j}$.
        \EndFor
        \Ensure $\mathcal{F}_{{\Theta}}^{r}(\cdot)$~and~$\mathcal{F}_{{\Theta}}^{d_j}(\cdot)$~, $j \in [1, J]$.
        \end{algorithmic}
    \end{algorithm}

\section{Experiments}
\label{sec:experiments}

\subsection{Experiment Settings}
\paragraph{Dataset and Splits}
We evaluate our self-supervised representation learning framework on QuickDraw 3.8M~\cite{xu2018sketchmate} dataset, which is a million-scale subset of Google QuickDraw dataset\footnote{\url{https://quickdraw.withgoogle.com/data}}~\cite{ha2018sketchrnn}. 
Our self-supervised training and associated validation are conducted on the training set~($3,105,000$ sketches) and validation set~($345,000$ sketches) of QuickDraw 3.8M.
After training, our self-supervised feature representations are tested on two sketch tasks (\ie, sketch retrieval and sketch recognition), and we 
extract features on query set~($34,500$ sketches) and gallery~($345,000$ sketches) of QuickDraw 3.8M. 
For sketch retrieval, we rank the gallery for each query sketch based on Euclidean distance, hoping the similar sketches ranked on the top. 
For sketch recognition, we \textcolor{black}{train a fully-connected layer as classifier} on gallery set, and calculate the recognition accuracy on query set. 
\paragraph{Evaluation Metric}
mAP~\cite{xu2018sketchmate} and classification accuracy~(``acc.'') are used as metrics for sketch retrieval and sketch recognition, respectively. In particular, for sketch retrieval, we calculate mAP over the top 1 and top 10 in retrieval ranking list, \ie, ``acc.@top1'' and ``mAP@top10''. 
\paragraph{CNN Implementation Details} 
The input size of our CNNs is $3 \times 224 \times 224$, with each brightness channel tiled equally. 
Plenty of CNN architectures can be utilized here, and 
for a fair comparison with our main competitor~\cite{gidaris2018unsupervised}, our CNNs are also implemented by AlexNet, with the output dimensionality is $4096D$.

\begin{figure}[!t]
	\centering
		\includegraphics[width=0.5\textwidth]{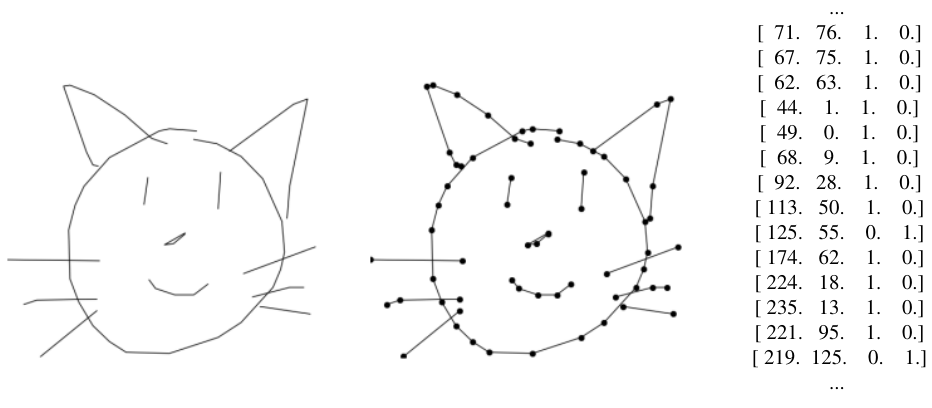}
	\caption{Stroke key point illustration of sketch. Each key point is denoted as a four-bit vector.}
	\label{fig:stroke_coordinate}
\end{figure}

\paragraph{TCN Implementation Details}
Based on statistic analysis on strokes, researchers have found most sketches of QuickDraw dataset have fewer than 100 strokes \cite{xu2018sketchmate}. Accordingly, the input array of our TCNs is normalized as $100 \times 4$ by truncating or padding. Each point is denoted as a four-dimensional vector, in which the first two number are \verb+x+ and \verb+y+ coordinates, and the last two bits describe pen state. Following the definition in~\cite{xu2018sketchmate}, the pen state is ``\verb+0 1+'' when the point is the stop point of one stroke. In remaining cases, the pen state is ``\verb+1 0+''~, as shown in Figure~\ref{fig:stroke_coordinate}.~\footnote{We experimentally found that if readuce the two-bit pen state as one-bit, similar results will be obtained. 
}
We implement a four-layer stacked TCN, where each layer has a series of $1$D convolution kernels with different sizes.
In particular, in the first layer of our TCN, $2$D convolution kernels are used to adaptive to our sketch coordinate input.
For the 2nd to 4th layers, we use $1$D convolution kernels. 
The implementation details of our TCN are reported in Table~\ref{table:TCN_structure}.
The output of each TCN layer is produced by ReLU activation and $1$D max pooling. 
The output dimensionality of our TCN is also $4096D$. 
During training, one fully-connected layer with batch normalization (BN)~\cite{Ioffe2015} and ReLU activation is used as the classifier for our TCN branch.

\paragraph{Selection of Deformations}
By observing a lot of sketch samples, we found that the most representative drawing styles of human mainly include horizontal compression (HC), centripetal compression (CC), vertical compression (VC), leftward compression (LC), and rightward compression (RC). Inspired by this observation, we empirically selected the corresponding deformations to conduct the pretext tasks. We found above selection leads to promising performance, and more deformations will be considered and compared in the future work.

\begin{table}
\caption{Implementation details of our TCN.  ``Conv2d\_Kx4'' and ``Conv1d\_K'' denote 2D convolution with kernel size of Kx4 and 1D convolution with kernel size of K, respectively. ``FC'' represents fully-connected layer.}
\resizebox{\columnwidth}{!}{
\begin{tabular}{c|c|c|c|c}
\hline
Input Shape & Operator & Channels & Kernel Size (K) & Stride \\
\hline 
\hline 
$100\times4$ & Conv2d\_Kx4 & 16 & 2,4,6,8,10,12,14,16,18,20 & 1 \\
$160$ & Conv1d\_K & 32 & 2,4,6,8,10,12,14,16,18,20 & 1 \\
$320$ & Conv1d\_K & 64 & 2,4,6,8,10,12,14,16,18,20 & 1 \\
$640$ & Conv1d\_K & 128 & 2,4,6,8,10,12,14,16,18,20 & 1 \\
$1280$ & FC & 4096 & - & - \\
$4096$ & FC & 345 & - & - \\
\hline
\end{tabular}}
\label{table:TCN_structure}
\end{table}

\paragraph{Other Implementation Details}
All our experiments are implemented in PyTorch~\footnote{https://pytorch.org/}~\cite{paszke2019pytorch}
, and run on a single 
GTX 1080 Ti GPU.
The detailed hardware and software configurations of our server are provided in Table~\ref{table:hardware}.
SGD optimizer (with initial learning rate $0.1$) and Adam optimizer (initial learning rate $0.001$) are used for CNNs and TCNs, respectively. 

\begin{table}[!t]
\small
\caption{Hardware and software details of our experimental environment.}
\label{table:hardware}
\begin{center}
\resizebox{\columnwidth}{!}{
\begin{tabular}{ l || c }
\hline

Hardware &  Configuration \\
\hline
\hline
CPU & two Intel(R) Xeon(R) CPUs (E5-2620 v3 @ 2.40GHz) \\
GPU & GEFORCE GTX 1080 Ti  (11GB RAM) \\
RAM & 128 GB \\
HD & solid state drive \\
\hline
\hline

Software &  Version \\
\hline
\hline
System &  Ubuntu 16.04 \\
Python &  3.6 \\
PyTorch &  0.4.1 \\
\hline
\end{tabular}}
\end{center}
\end{table}

\paragraph{Competitors}
We compare our self-supervised representation approach with several the state-of-the-art deep unsupervised representation techniques, including self-supervised~(\textbf{RotNet}~\cite{gidaris2018unsupervised}, \textbf{Jigsaw}~\cite{noroozi2016unsupervised}), clustering-based~(\textbf{Deep Clustering}~\cite{caron2018deep}), generative adversarial network based~(\textbf{DCGAN}~\cite{radford2015unsupervised}), and auto-encoder based~(\textbf{Variational Auto-Encoder}~\cite{kingma2013auto}) approaches. 
For a fair comparison, we evaluate all competitors based on the same backbone network if applicable. 
Moreover, in order to evaluate the viewpoint~\cite{kolesnikov2019revisiting} that  standard network architecture design recipes do not necessarily translate from fully-supervised setting to self-supervised setting, we also implement some baselines by replace our feature extractor with RNN.

\begin{table}[!t]
\small

\caption{Comparison on retrieval task with state-of-the-art deep learning based unsupervised methods. ``R'' denotes ``rotation''. ``\&'' means that two deformations are applied simultaneously. The $1^{st}$/$2^{nd}$~best results on column basis are indicated in \textcolor{red}{red}/\textcolor{blue}{blue}.}
\label{table:retrieval_comparison_with_baselines}
\begin{center}
\resizebox{\columnwidth}{!}{
\begin{tabular}{ l || c | c }
\hline
Unsupervised Baselines & acc.@top1 & mAP@top10\\
\hline
\hline
DCGAN~\cite{radford2015unsupervised} & 0.1695  & 0.2239 \\
Auto-Encoder~\cite{kingma2013auto} & 0.0976 & 0.1539 \\
Jigsaw~\cite{noroozi2016unsupervised} & 0.0803 & 0.1270 \\
Deep Clustering~\cite{caron2018deep} & 0.1787 & 0.2396 \\
R+CNN (RotNet) ~\cite{gidaris2018unsupervised} & \textcolor{blue}{0.4706} & \textcolor{blue}{0.5166}  \\
\hline
\hline
\tabincell{l}{RNN Self-Sup. Baselines We Designed\\ \{\textit{pretext task}\}+\{\textit{feature extractor}\}} & acc.@top1 & mAP@top10\\
\hline
\hline
\{R\}+\{RNN\} & 0.0234 & 0.0533 \\
\{HC\}+\{RNN\} & 0.0218 & 0.0488 \\
\{VC\}+\{RNN\} &  0.0226 & 0.0507 \\
\{CC\}+\{RNN\} & 0.0125 & 0.0312 \\
\{VC\&LC\}+\{RNN\} & 0.0210 & 0.0481 \\
\{VC\&RC\}+\{RNN\} & 0.0186 &0.0446 \\
\hline
\hline
\tabincell{l}{Our Full Model\\ \{\textit{pretext task}\}+\{\textit{feature extractor}\}} & acc.@top1 & mAP@top10\\
\hline
\hline
\tabincell{l}{\{R,HC,VC,CC,VC\&LC,VC\&RC\}\\+\{CNN,TCN\}} & \textcolor{red}{0.5024} &  \textcolor{red}{0.5447} \\
\hline
\end{tabular}}
\end{center}

\end{table}

\begin{figure*}[!t]
	\centering
		\includegraphics[width=\textwidth]{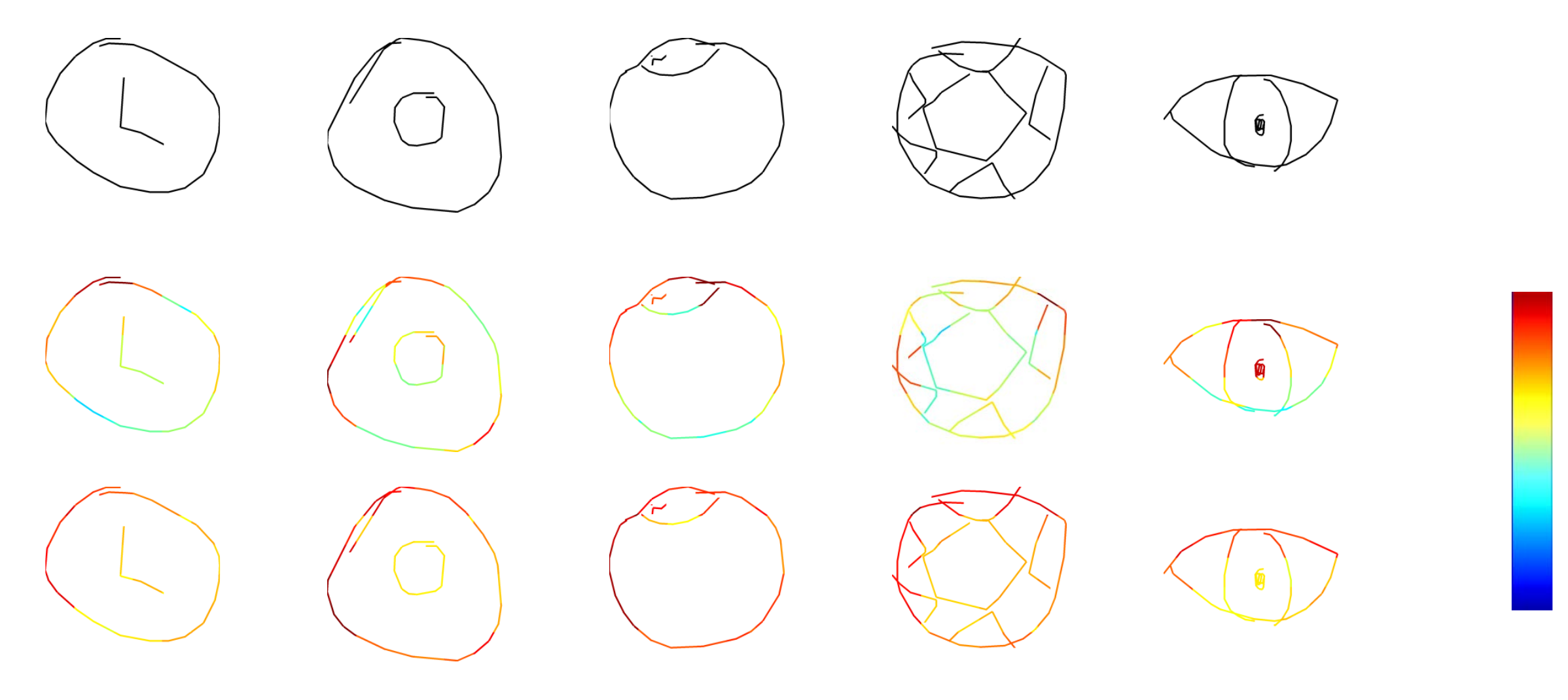}
	\caption{Attention map visualization (clock, donut, blueberry, soccer ball, eye). Color bar ranging from blue to red denotes activated values 0 to 1. Original sketches are in the top row. Middle and bottom rows are obtained by RotNet and our full model, respectively. Best viewed in color.}
	\label{fig:attention_map}
\end{figure*}

\subsection{Results and Discussions}
\paragraph{Evaluation on Sketch Retrieval}
We evaluate our self-supervised learned features on sketch retrieval, by comparing with the features obtained by the state-of-the-art unsupervised representation methods.
Retrieval results ``acc.@top1'' and ``mAP@top10'' are reported in Table~\ref{table:retrieval_comparison_with_baselines}, where following observations can be made: (i) Except for RotNet~\cite{gidaris2018unsupervised}, all other baselines fail to work well on sketch unsupervised representation learning due to the unique challenges of sketch. 
Particularly, Jigsaw~\cite{noroozi2016unsupervised} obtained low retrieval accuracy due to that sketch patches are too abstract to recognize. (ii)
RotNet~\cite{gidaris2018unsupervised} outperforms other baselines by a clear margin, showing us the effectiveness of ``image-level'' self-supervised pretext tasks for abstract sketch. (iii) Our proposed method obtains better retrieval results over all the baselines listed in Table~\ref{table:retrieval_comparison_with_baselines}. (iv) It is interesting that RNN extractor obtains unsatisfactory performance with a number of self-supervised pretext tasks, whilst RNN networks have achieved the state-of-the-art performance~\cite{xu2018sketchmate} in supervised settings. This confirms that the network designing recipe under fully supervised settings can not be directly transfered to self-supervised setting, which has been demonstrated in~\cite{kolesnikov2019revisiting}. This also demonstrates the necessity of our sketch-specific architecture design in self-supervised feature extraction setting.
\begin{table}[!t]
\small
\caption{Sketch retrieval ablation study on our proposed self-supervised representation learning framework. ``\&'' means that two deformations are applied simultaneously. The $1^{st}$/$2^{nd}$~best results on column basis are indicated in \textcolor{red}{red}/\textcolor{blue}{blue}.}
\label{table:retrieval_deformation_ablation}
\begin{center}
\resizebox{\columnwidth}{!}{
\begin{tabular}{ l || c | c }
\hline

\tabincell{l}{Abbr.\\ \{\textit{pretext task}\}+\{\textit{feature extractor}\}} & acc.@top1  & mAP@top10\\
\hline
\hline
\{HC\}+\{CNN\} &  0.1932 & 0.2556 \\
\{HC\}+\{TCN\} &  0.1229 & 0.1756  \\
\{HC\}+\{CNN,TCN\} &  0.2388 & 0.2994 \\
\{VC\}+\{CNN\} &  0.1800 &  0.2433 \\
\{VC\}+\{TCN\} & 0.1468 &  0.2008 \\
\{VC\}+\{CNN,TCN\} &  0.2435 & 0.3025 \\
\{CC\}+\{CNN\} &  0.2555 & 0.3159 \\
\{CC\}+\{TCN\} &  0.1489 &  0.2048 \\
\{CC\}+\{CNN,TCN\} &  0.2876 & 0.3428 \\
\{VC\&LC\}+\{CNN\} &  0.2459 & 0.3053 \\
\{VC\&LC\}+\{TCN\} & 0.2003 &  0.2580 \\
\{VC\&LC\}+\{CNN,TCN\} &  0.2574 & 0.3132 \\
\{VC\&RC\}+\{CNN\} &  0.2265 & 0.2879 \\
\{VC\&RC\}+\{TCN\} &  0.1870 &  0.2427 \\
\{VC\&RC\}+\{CNN,TCN\} &  0.2367 & 0.2931 \\
\hline
\hline
\{HC,VC\}+\{CNN,TCN\} &  0.2842 & 0.3404 \\
\{HC,VC,CC\}+\{CNN,TCN\} &  0.3060 & 0.3582 \\
\{HC,VC,CC,VC\&LC\}+\{CNN,TCN\} &  \textcolor{blue}{0.3060} & \textcolor{blue}{0.3685} \\
\tabincell{l}{\{HC,VC,CC,VC\&LC,VC\&RC\}\\+\{CNN,TCN\}} & \textcolor{red}{0.3180} & \textcolor{red}{0.3718} \\
\hline
\end{tabular}}
\end{center}

\end{table}
\begin{table}[!t]
\small
\caption{Sketch retrieval ablation study on the contribution of dual-branch CNN-TCN to rotation-based self-supervised learning. ``R'' denotes ``rotation''. The $1^{st}$/$2^{nd}$~best results on column basis are indicated in \textcolor{red}{red}/\textcolor{blue}{blue}.}
\label{table:rotation_ablation}
\begin{center}
\resizebox{\columnwidth}{!}{
\begin{tabular}{ l || c | c }
\hline

\tabincell{l}{Abbr.\\ \{\textit{pretext task}\}+\{\textit{feature extractor}\}} &  acc.@top1  & mAP@top10\\
\hline
\hline
\{R\}+\{CNN\} (RotNet)~\cite{gidaris2018unsupervised} &  \blue{0.4706} & \blue{0.5166}  \\
\{R\}+\{TCN\} & 0.3072 & 0.3639 \\
\{R\}+\{CNN,TCN\} &   \red{0.4932} & \red{0.5360} \\
\hline
\end{tabular}}
\end{center}
\end{table}

Although RotNet is the strongest baseline to ours, 
its rotation-based pretext task fails to work well on centrosymmetric sketches, \eg, donut, soccer ball.
Intuitively, given a centrosymmetric sketch, visual variation caused by rotation is limited and difficult to captured even for human eye.
We visualize attention maps for some centrosymmetric sketches in Figure~\ref{fig:attention_map}, where middle and bottom rows are obtained by RotNet and our full model, respectively.
Based on our color bar, we observe that compared with the attention maps in the middle row, ours have larger activated values. 
This means that our proposed model works more sensitively to the strokes of centrosymmetric sketches. 

Moreover, we also conduct some ablation studies on retrieval to evaluate the contributions of our deformation-based pretext tasks and CNN-TCN architecture, by combining different pretext tasks and feature extractors within our proposed self-supervised framework.
From Table~\ref{table:retrieval_deformation_ablation}, we observe that: (i) Given a deformation-based pretext task, our dual-branch CNN-TCN brings performance improvement over CNN and TCN. (ii) Based on our CNN-TCN feature extraction, with more kinds of deformation-based pretext tasks involved, better performance will be achieved. 

To further demonstrate the generality of our CNN-TCN architecture on image-level self-supervised pretext tasks, we also implement ablation study to evaluate whether CNN-TCN architecture can improve rotation-based self-supervised method, \ie, RotNet.  
Table~\ref{table:rotation_ablation} shows that CNN-TCN extractor brings performance improvement for rotation-based self-supervised learning, and outperforms both single-branch CNN and TCN. This phenomenon also illustrates that CNN and TCN could produce complementary features in sketch self-supervised learning setting. 

\paragraph{Evaluation on Sketch Recognition}
We also evaluate our self-supervised learned features on sketch recognition task.  We train our model on QuickDraw 3.8M training set, and extract features for its gallery set and query set. Then, we use gallery features and the associated ground-truth labels to train a linear classifier. Classification accuracy is calculated on QuickDraw 3.8M query set. Similar operations are performed for our competitors. For a fair comparison, we keep the classifier configuration the same for all our classification experiments. 

The following observations can be obtained from Table~\ref{table:classification_comparison_with_baselines}: (i) For sketch recognition, our model and its variant outperform the state-of-the-art unsupervised competitors by a large margin ($0.5652$ vs. $0.5149$), demonstrating the superiority of our sketch-specific design. 
(ii) When stroke deformation based self-supervised signals are added, $1.79$ percent improvement on recognition accuracy is obtained.
(iii) The performance gap between the state-of-the-art supervised sketch recognition model Sketch-a-Net~\cite{yu2017sketch} and ours is narrow ($0.6871$ vs. $0.5652$).

\begin{table}[t!]
\tiny
\caption{Comparison on sketch recognition with the state-of-the-art deep learning based unsupervised methods.``R'' denotes ``rotation''. ``\&'' means that two deformations are applied simultaneously. The $1^{st}$/$2^{nd}$~best results are indicated in \textcolor{red}{red}/\textcolor{blue}{blue}.}
\label{table:classification_comparison_with_baselines}
\begin{center}
\resizebox{0.4\textwidth}{!}{
\begin{tabular}{ l || c }
\hline
Unsupervised Baselines & acc.\\
\hline
\hline
DCGAN~\cite{radford2015unsupervised} & 0.1057 \\
Auto-Encoder~\cite{kingma2013auto} & 0.1856\\
Jigsaw ~\cite{noroozi2016unsupervised} &  0.2894 \\
Deep Clustering~\cite{caron2018deep} & 0.0764 \\
R+CNN (RotNet)~\cite{gidaris2018unsupervised} &  0.5149  \\
\hline
\hline
\tabincell{l}{Our Method Abbr.\\ \{\textit{pretext task}\}+\{\textit{feature extractor}\}} & acc.\\
\hline
\hline
\{R\}+\{CNN,TCN\} &   \textcolor{blue}{0.5473} \\
\tabincell{l}{\{R, HC,VC,CC,VC\&HLE,VC\&HRE\}\\+\{CNN,TCN\}} &  \textcolor{red}{ 0.5652} \\
\hline
\hline
Supervised Methods & acc. \\
\hline 
\hline
Sketch-a-Net~\cite{yu2017sketch} & 0.6871 \\
\hline
\end{tabular}}
\end{center}
\end{table}
\section{Conclusion}
\label{sec:conclusion}
In this paper, we propose the novel problem of self-supervised representation learning for  free-hand sketches, and contribute the first deep network based framework to solve this challenging problem.
In particular, by recognizing the intrinsic traits of sketches, we propose a set of sketch-specific self-supervised pretext tasks, and a dual-branch TCN-CNN architecture serving as feature extractor.
We evaluate our self-supervised representation features on two tasks of sketch retrieval and sketch recognition. 
Our extensive experiments on million-scale sketches demonstrate that our proposed self-supervised representation method outperforms the state-of-the-art unsupervised competitors, and significantly narrows the gap with supervised representation learning on sketches. 

We sincerely hope our work can motivate more self-supervised representation learning in the sketch research community.


\section{Future Work}
\label{sec:future_work}
As the aforementioned, free-hand sketch has its domian-unique technical challenges, since it is essentially different to natual photo.
Therefore, designing sketch-specific pretext tasks for free-hand sketch oriented self-supervised deep learning is significant.
In particular, in future work, we will try to design sketch-specific pretext tasks from fine-grained perspective, involing more stroke-level analysis.


{\small
\bibliographystyle{IEEEtran}
\bibliography{egbib}
}

\end{document}